# Syntax-Aware Language Modeling with Recurrent Neural Networks

Duncan Blythe, Alan Akbik and Roland Vollgraf
Zalando Research, Zalando SE
firstname.lastname@zalando.de

March 2, 2018

## Abstract

Neural language models (LMs) are typically trained using only *lexical* features, such as surface forms of words. In this paper, we argue this deprives the LM of crucial *syntactic* signals that can be detected at high confidence using existing parsers. We present a simple but highly effective approach for training neural LMs using both lexical and syntactic information, and a novel approach for applying such LMs to unparsed text using sequential Monte Carlo sampling. In experiments on a range of corpora and corpus sizes, we show our approach consistently outperforms standard lexical LMs in character-level language modeling; on the other hand, for word-level models the models are on a par with standard language models. These results indicate potential for expanding LMs beyond lexical surface features to higher-level NLP features for character-level models.

## 1 Introduction

Language modeling (LM) is the project of estimating the probability of sequences of words as they occur in natural language [3]. LM is used either to (1) *score* a word sequence in order to distinguish ungrammatical or nonsensical from well-formed sentences [23], or (2) to *generate* word sequences, useful for e.g. auto-generating image captions [27], decoding target language translations in neural machine translation [26], and auto-completion of sentences [18].
**Neural language models.** Advances in recurrent neural networks (RNNs) [11, 24] and sequence-to-sequence learning [25] have given rise to renewed interest in language modeling, prompting a range of work in *neural language models* [10, 5, 13]. These approaches have been shown to far outperform earlier $N$-Gram based models [12] due to the ability of RNNs to flexibly encode long-term dependencies through a hidden-state.
**Explicitly modeling syntax.** Despite this progress, current LMs typically use only lexical features (i.e. textual tokens or characters) as features over which to learn a probabilistic model of text. It may be argued that this deprives the language model of crucial signals such as the shallow and deep syntactic structure of sentences which can be detected at relatively high confidence using existing part-of-speech (PoS) taggers or deep syntactic parsers [16]. Such features are theoretically very useful for language modelling, as they make explicit distinctions between word types (for instance verbs vs. nouns), or between different grammatical functions of a word (grammatical subject vs. object).
**Syntax-aware language models.** In this paper, we investigate the hypothesis that LMs improve if we include explicit word-level syntactic information in the training process, allowing them to better score and generate more plausible and grammatically correct sentences. We propose including syntactic features during training by first parsing all sentences in the corpus, and then simply rendering word-level syntactic information as token prefixes. See Figure 1 for an illustration. A core advantage of this approach is that by training an RNN language model over such sequences, the model will be encouraged to write both syntactic and lexical information to its hidden-state. We refer to this proposed LM as *syntax-aware language model* (SALM).

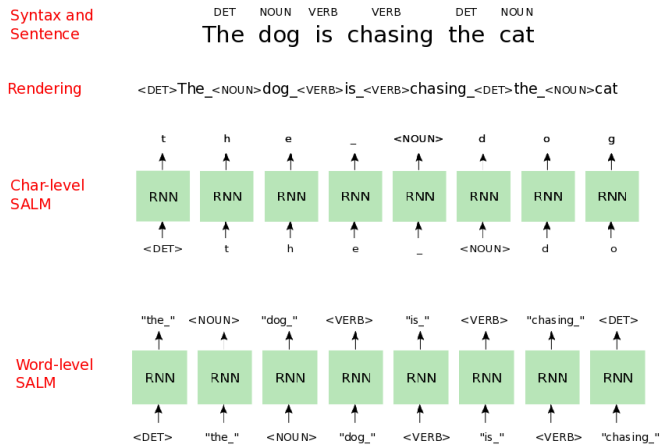

Figure 1: Training process of the proposed syntax-aware language models. We parse each sentence (1) and generate sequences in which syntactic information is rendered as token-prefixes (2). We then use these generated sequences to train a character-based LSTM language model (3). Token-prefixes are always treated as one character when training the LSTM.

**Applicability of SALMs.** Although training such a model over annotated text is relatively straightforward using standard techniques for training RNNs, a crucial question is how to apply such an LM to plain text (i.e. text for which lexical, but no syntactic features are available). This applicability is a core requirement in many applications such as scoring unparsed text and predicting probable continuations of a text fragment. Most importantly, it is necessary to allow for comparative evaluation with standard LMs that use only lexical features. To address this issue, we propose a sequential Monte Carlo approach for leveraging our LM to infer the probable syntactic structures of a plain sentence or fragment, and then performing scoring or text generation based on this inferred structure. We refer to our proposed algorithm as SYNSIR.

**Contributions.** To summarize, this paper proposes syntax-aware language models (SALMs) and addresses the questions of (1) how to include syntactic information in model training, and (2) how to apply models trained over syntactic features to unparsed text. We present a set of experiments over various corpora and sample sizes. The experiments show that our approach consistently outperforms standard neural language models in terms of model perplexity for character level models; interestingly, however, on the word-level, the proposed models are almost exactly on a par with the baseline language model. We believe that these experiments point to great potential for including a richer set of linguistic features into character level language modeling, while retaining full applicability to unparsed text.

## 2 Previous Work

### 2.1 Neural Language Modeling with LSTMs

A commonly used method for training neural language models utilizes a long-short term memory (LSTM) RNN architecture, trained to predict the next character or word from the previous characters or words [10]. In this paper, we also adopt this architecture. Refer to Figure 1 for an illustration of the prediction problem: tokens are passed in sequence as inputs to the LSTM cells (illustrated below each cell) and trained to predict the next token (illustrated above each cell).

We adopt a character-level representation, in which a sentence is a sequence of characters $\mathbf{z}_{0:t} = z_0, z_1, \ldots, z_t$. The goal of modeling is to obtain a good distribution over such sequences $p(\mathbf{z}_{0:t})$ reflecting natural language production [5]. By training a language model, we learn $P(z_t|z_0, \ldots, z_{t-1})$. The joint distribution over entire sentences may then be decomposed as a product of the predictive distribution over characters given the first $i$ letters:

$$P(\mathbf{z}_{0:i}) = \prod_i P(z_i|\mathbf{z}_{0:i-1}) \quad (1)$$

Here one models the probability of character $z_t$ as a function of the output of an LSTM:

$$P(z_{t+1}) = \text{softmax}(v_t)[z_{t+1}] \quad (2)$$

In vanilla LSTM LMs [10] the state, $m_t$, is the result of a non-linear recurrence update which conditions on the previous state and the latest character of the sequence:

$$u_t = W_e[z_t] \quad (3)$$

$$\begin{bmatrix} i_t \\ j_t \\ f_t \\ o_t \end{bmatrix}^\top = W \begin{bmatrix} u_t \\ m_{t-1} \end{bmatrix} + b \quad (4)$$

$$c_t = \sigma(f_t) \circ c_{t-1} + \sigma(i_t) \circ h(j_t) \quad (5)$$

$$v_t = W_o\, m_t + b_o \quad (6)$$

The first line denotes that the input $z_t$ is first embedded linearly before being input to the LSTM cells.

Here, the nonlinearities $\sigma$ and $h$ are given by the sigmoid and hyperbolic tangent respectively. More succinctly we may summarize this nonlinear recurrence as:

$$m_t, c_t = f(x_t, m_{t-1}, c_{t-1}) \quad (7)$$

The motivation of the various variables involved in the update equations has been explained in detail elsewhere [11]. Briefly, $c_t$ corresponds to a "memory" for storing past sequence information. This memory may be written to (left hand term of Equation (5)) and also erased (right hand term of Equation (5)). $i_t, j_t, f_t$ are variables which determine by how much and whether the memory is updated or erased, which are calculated linearly from the current input $z_t$ and previous output $m_{t-1}$ (Equation (4)). The output gate $o_t$ determines whether the network returns a non-zero output (Equation (6)).

The language model we use in this paper uses several embellishments, on top of this standard structure. Firstly, in training we apply dropout to the input and output of the LSTM [28]:

$$u_t = \text{drop}_{p_d}(W_e[z_t]) \tag{8}$$

$$P(z_{t+1}) = \text{softmax}(\text{drop}_{p_d}(m_t))[z_{t+1}] \tag{9}$$

Secondly, we stack two LSTM layers [12], with a number, $n_h$, of hidden units and taking the projected output of the second as input to the softmax layer:

$$m'_t, c'_t = f_1(x_t, m'_{t-1}, c'_{t-1}) \tag{10}$$

$$m_t, c_t = f_2(m'_t, m_{t-1}, c_{t-1}) \tag{11}$$

Here $f_1$ is the non-linear recurrence of the first layer and $f_2$ the recurrence of the second layer as in Equation (7).

## 2.2 Related Work on Parsing as Language Modeling

[5] recently showed that parsing and language modeling may be unified by training over character string renderings of constituency parsed sentences. They demonstrated that a joint model $P(\mathbf{x}, \mathbf{y})$ over sentences and parses may be used to score the output of an $n$-best parser which returns several possible deep syntactic interpretations for a sentence. Using their LM, they selected the most probable parses from these choices and so created a system that more reliably predicts parse trees for sentences, achieving competitive scores on a standard Penn Treebank [17] benchmark dataset.

However, while their work illustrates the benefits of including syntactic features into a character-based neural network model, their work was limited in scope to scoring fully predicted parse trees. They argued that this approach cannot be used to either score or parse an unparsed sentence (i.e. without requiring an existing $n$-best parser) since a brute-force approach of first generating all conceivable syntactic interpretations and then scoring them is impossible on the grounds of computational complexity. Moreover, due to the nested structure of their syntactic markup, applying their model to word-completion is non-trivial and was not attempted in the paper.

Our work follows a similar motivation, but instead of a nested tree structure we render word-level syntactic information in the form of a simpler token-prefix markup. Additionally, we show how to use approximate sampling methods to apply the trained model to unparsed text, for instance to predict probable subsequent words given a text fragment. This method allows us to add syntactic information during the training of the LM, but retain full applicability to unparsed text.

[22] use a graphical model approach to modeling language over tree structures; their approach differs from ours in requiring complete parse trees to generate language, and thus cannot be applied straightforwardly in an online fashion to, for instance, sentence continuation. In addition their approach does not benefit from the powerful LSTM model studied in this paper. [9] use a neural language model, an equivalence class approach and an EM algorithm to estimating unseen annotations on new data. Their approach differs from ours in three respects. In our work "equivalence classes" of vocabulary are subsumed into the embeddings learnt by the LSTM. In contrast to our RNN approach the authors use a less powerful feedforward neural network structure. Finally the estimates of annotations produced by their proposed EM algorithm cannot be easily updated online, to incorporate incoming words in an incomplete sentence-fragment, unlike the algorithm we present below. Finally [14] use particle filtering in a sequential manner given a PCFG model to predict the memory load of human participants in a linguistic task. Our work coincides in that both use sequential Monte-Carlo techniques. However their work aims at understanding human linguistic comprehension rather than improving language models, and does not incorporate a powerful RNN.

# 3 Syntax-Aware Language Models

We now present our proposed syntax-aware neural language model (SALM). We first illustrate how we render syntax sequentially and how we train the LSTM over this data (Subsection 3.1). We then discuss the crucial question of how to apply the model to unparsed sentences that lack this markup for the purposes of scoring and word completion (Subsection 3.2), proposing the novel SYNSIR algorithm.

## 3.1 Training Syntax-Aware Language Models

We train our neural language model over sequences that contain both lexical and syntactic features. To generate these sequences, we first parse each sentence in the training corpus, and then produce a modified sequence that includes syntactic prefixes to each token. As Figure 1 shows, we render sentences in three variations that reflect different types of syntactic information:

To simplify the exposition and experiments, we apply ouf SALMs to shallow-syntactic part-of-speech tags (PoS) [19]; in principle, however the framework may be applied to any type of word-level syntactic annotations. PoS tags are a set of word-level syntactic tags that generalize across languages. Due to this multilingual generalization, no language-specific word-level syntax is modeled. These tags only make general distinctions between word classes such as verbs, nouns, determiners and adjectives.

**Training the LSTM.** Formally, the sentence and parse may be considered as a single sequence $\mathbf{z}$ consisting of alternating words $x_i$ and their syntax prefixes $y_i$. In the character LM case, the prefixes are considered a single character and in the word LM case, a single word.

$$\begin{aligned} \mathbf{z} &= z_1, z_2, \ldots, z_t \\ &= y_1, x_1, y_2, x_2, \ldots, y_n, x_n \end{aligned} \quad (12)$$

We train the LSTM on these sequences. Due to the small number of distinct syntax prefixes, this leads to only a small increase in the size of the vocabulary. As in standard RNN LMs, each prefix is encoded by a one-hot vector. Given this representation, the learnt model gives us a joint distribution over syntax prefixes and words by factorizing:

$$P(\mathbf{x}, \mathbf{y}) = \prod_i P(x_i | \mathbf{x}_{0:i-1}, \mathbf{y}_{0:i}) \cdot P(y_i | \mathbf{x}_{0:i-1}, \mathbf{y}_{0:i-1}) \quad (13)$$

In this way predictions for the next word $x_i$ are always conditioned on the syntax prefix $y_i$ for that word. This conditioning reduces the space of possibilities for the next word, allowing the network to learn a better model. For example, knowing that the next word has a noun-indicating syntax prefix implies that all words which only appear as verbs are excluded.

Note that this approach is equally applicable to character and word level LMs, with the proviso that, in the character case, words $x_i$ consist of short sequence of characters.

## 3.2 Application to Unparsed Text

As previously mentioned, a crucial question is how to apply the trained SALM to unparsed text. While the SALM is trained on alternating sequences of syntax prefixes and words, a plain-text fragment is given by a plain-text sequence without syntax prefixes.

Naïvely, one might attempt to simply parse the text to which the SALM is applied using the same parser that was used to train the model. However, this approach is impractical in many typical

LM tasks such as word prediction or sentence completion that operate on incomplete text or word fragments. An example might be a query auto-complete function in a Web search engine that suggests queries based on the character sequence a user has typed so far. Since parsers are generally trained on and optimized for complete sentences with fully typed-out words they cannot reliably handle such scenarios. For this reason we need to leverage the probabilistic information learnt by the SALM which, by contrast, treats all presented text as fragments due to the sequential nature of the training.

Mathematically speaking, applying the model to unparsed data is tantamount to calculating $P(x_i|\mathbf{x}_{0:i-1})$ from the trained LSTM—i.e. the probability of the next word given the prior words. However the SALM training gives us $P(x_i|\mathbf{y}_{0:i}\mathbf{x}_{0:i-1})$ where the $\mathbf{y}_{0:i}$ are unknown. Formally we may obtain $P(x_i|\mathbf{x}_{0:i-1})$ by marginalizing over $\mathbf{y}_{0:i+1}$:

$$P(x_i|\mathbf{x}_{0:i-1}) \tag{14}$$

$$= \sum_{\mathbf{y}_{0:i}} P(x_i, \mathbf{y}_{0:i}|\mathbf{x}_{0:i-1}) \tag{15}$$

$$= \sum_{\mathbf{y}_{0:i}} P(x_i, y_i|\mathbf{y}_{0:i-1}, \mathbf{x}_{0:i-1}) \tag{16}$$

$$\times P(\mathbf{y}_{0:i-1}|\mathbf{x}_{0:i-1}) \tag{17}$$

However, actually computing this marginalization is not possible, for large $i$, due to the number of possible values of the syntax prefix sequence $\mathbf{y}_{0:i-1}$ participating in the sum. Instead we seek a discrete approximation to Equation (17), by approximating the expectation over $P(\mathbf{y}_{0:i-1}|\mathbf{x}_{0:i-1})$ using sequential importance resampling (SIR) [8].

SIR is a technique to obtain an approximation to $P(\mathbf{y}_{0:i-1}|\mathbf{x}_{0:i-1})$ which may be sequentially updated to yield an approximation to $P(\mathbf{y}_{0:i}|\mathbf{x}_{0:i})$ as new data is received; here one understands the $\mathbf{x}_{0:i}$ as an observed sequence and $\mathbf{y}_{0:i}$ an unobserved or "hidden" sequence influencing how the observations arise. In our case the observations are the words in the sentence and the hidden values are the syntax.

SIR assumes one specifies a transition kernel $\pi(y_i|\mathbf{y}_{0:i-1}, \mathbf{x}_{0:i})$ and that we have "particles" $\mathbf{y}_{0:i-1}^j$ for $j = 1, \ldots, n$, and weights $w_j$ so that the posterior up until $i-1$ is approximated by

$$P(\mathbf{y}_{0:i-1}|\mathbf{x}_{0:i-1}) \approx \sum_j w_{i-1}^j \delta(\mathbf{y}_{0:i-1} - \mathbf{y}_{0:i-1}^j)$$

Using this kernel, new hidden values $y_i^j$ are sampled:

$$y_i^j \sim \pi(y_i|\mathbf{y}_{0:i-1}^j, \mathbf{x}_{0:i})$$

New weights are calculated according to:

$$w_i^j = w_{i-1}^j \cdot \frac{P(x_i|[y_i^j, \mathbf{y}_{0:i-1}^j], \mathbf{x}_{0:i-1}) P(y_i^j|\mathbf{y}_{0:i-1}^j, \mathbf{x}_{0:i-1})}{\pi(y_i|\mathbf{y}_{0:i-1}^j, \mathbf{x}_{0:i})}$$

$$w_i^j = \frac{w_i^j}{\sum_j w_i^j}$$

This update step guarantees for large $n$ that:

$$P(\mathbf{y}_{0:i}|\mathbf{x}_{0:i}) \approx \sum_j w_i^j \delta(\mathbf{y}_{0:i} - \mathbf{y}_{0:i}^j)$$

As transition kernel we use the learnt prior distribution over the next annotation given the previous annotations and words:

$$\pi(y_i|\mathbf{y}_{0:i-1}, \mathbf{x}_{0:i}) = p(y_i|\mathbf{x}_{0:i-1}\mathbf{y}_{0:i-1})$$

We also experimented with using an 'optimal' transition kernel [2]; this kernel is given by the posterior over hidden states given the new observation $x_i$:

$$\pi(y_i|\mathbf{y}_{0:i-1}, \mathbf{x}_{0:i}) = p(y_i|\mathbf{x}_{0:i}\mathbf{y}_{0:i-1})$$

However we found the performance gains obtained with this method were balanced by the far greater computational demand of performing the marginalization necessary for calculating the optimal kernel. This becomes especially problematic for large vocabulary sizes.

A practical problem is that many of the $w_i^j$ become very small as plausible initial hidden sequences become unlikely in light of new observations. For this reason, in SIR whenever the number of weights with $w_i^j < \gamma$ falls below a threshold $n_{thresh}$, one resamples and reassigns weights:

$$\mathbf{y}_{0:i}^j \sim \sum_j w_i^j \delta(\mathbf{y}_{0:i} - \mathbf{y}_{0:i}^j)$$

$$w_i^j = 1/n$$

This ensures that the particles obtained do not have negligible posterior probability. To simplify the algorithm and to eliminate hyper-parameters we resampled after every word. In practice we found, however, that this repeated resampling did not greatly affect performance.

After obtaining samples $\mathbf{y}_{0:i-1}$ and normalized weights $w_0, \ldots w_{i-1}$ using SIR they may be used to approximate the likelihood $P(x_i|\mathbf{x}_{0:i-1})$ in the following way:

$$\begin{align}
& P(x_i|\mathbf{x}_{0:i-1}) \tag{18}\\
=\ & \sum_{\mathbf{y}_{0:i}} P(x_i, \mathbf{y}_{0:i}|\mathbf{x}_{0:i-1}) \tag{19}\\
=\ & \sum_{\mathbf{y}_{0:i}} P(x_i|\mathbf{y}_{0:i}, \mathbf{x}_{0:i-1}) \\
& \times P(\mathbf{y}_{0:i}|\mathbf{x}_{0:i-1}) \tag{20}\\
\approx\ & \sum_j w_j \cdot P(x_i|\mathbf{y}_{0:i}^j, \mathbf{x}_{0:i-1}) \tag{21}\\
:=\ & \widehat{P}(x_i|\mathbf{x}_{0:i-1}) \tag{22}
\end{align}$$

Our SIR algorithm for SALMs which we call SYNSIR is detailed in Algorithm 1. The algorithm takes a sentence and a trained SALM model as input and outputs an SIR approximation to the posterior over syntax prefixes given the sentence as well as estimates of the predictive distribution $\widehat{P}(x_i|\mathbf{x}_{0:i-1})$.

## 4  Experiments

We conduct a comparative evaluation of standard LSTM-LM against the proposed SALM. We wish to determine if and how language modeling, as measured in model perplexity, improves using syntax-awareness; note however, that the aim of the approach is not to perform parsing, but rather to use syntactic awareness to improve language modeling. For this reason, we do not evaluate on standard parsing/ part of speech tasks, but rather on plain-text datasets.

**Algorithm 1** SYNSIR: Sequential Importance Resampling for SALMs
---

**function** SYNSIR($P, \mathbf{x}_{0:n}, m$)
    $P$: $SALM$ model, $\mathbf{x}_{0:n}$: sentence, $\gamma$: particle degeneracy constant, $n_{thresh}$: number of small particles allowed
    **for** i=0...n **do**
        **for** j=0...m **do**
            Sample $y_i^j \sim P(y|\mathbf{x}_{0:i}, \mathbf{y}_{0:i-1}^j)$
            $\alpha_j = P(x_i|\mathbf{x}_{0:i-1}, \mathbf{y}_{0:i}^j)$
            $\mathbf{y}_{0:i}^j = [\mathbf{y}_{0:i-1}^j, y_i^j]$
        **end for**
        $w_j = \alpha_j / \sum_j \alpha_j$
        $\widehat{P}(x_i|\mathbf{x}_{0:i-1}) = \frac{1}{m}\sum_{i=1}^m \alpha_i$
        Resample $\mathbf{y}_{0:i}^j$ with replacement ($j = 1 \ldots m$) from $\mathbf{w}$
    **end for**
    **return** $\mathbf{y}^i, \widehat{P}(x_i|\mathbf{x}_{0:i-1})$
**end function**

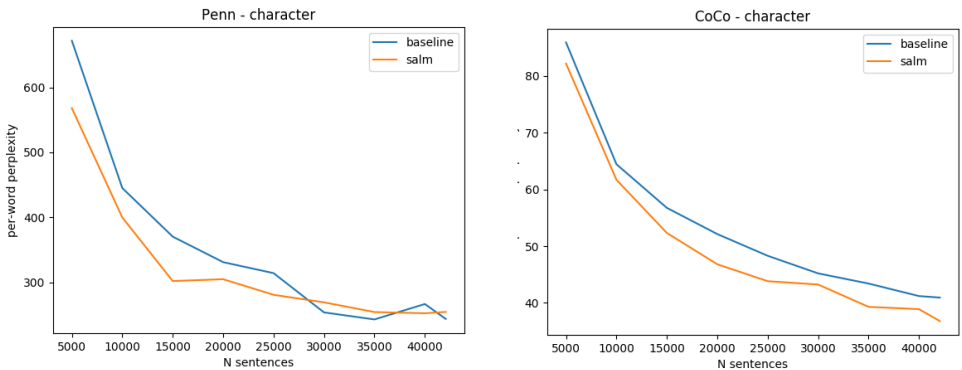

Figure 2: Results of character level experiments

Our experiments compare SALMs to a purely lexical baseline LSTM-LM. This is because LSTMs have been shown to exhibit state-of-the-art performance in terms of perplexity [12]. However, our approach may equally be applied to other effective recurrent neural network architectures, such as those based on gated recurrent units [6], and in general to any language model based on modeling $P(x_{t+1}|x_0, \ldots, x_t)$; this includes traditional $N$-gram models. We defer a broader comparison to later work.

## 4.1 Experimental Setup

**Datasets.** In order to evaluate the general capability of SALMs to model varied language, we utilize two English-language corpora that each have distinct lexico-syntactic writing styles. These are:

- PENN treebank dataset [17]. This dataset comprises approximately 40,000 English sentences taken from a database of news articles, also supplied with syntactic annotations (PoS, deep syntactic parse). In this experiment we use only the plain text.

- COCO dataset [15]. The first dataset comprises approximately 600,000 captions of pictures of everyday objects in context. These captions were obtained through crowdsourcing, are gener-ally short and declarative, and often lack a main verb ("a man with his dog at the beach"). We use only the first 44,098 sentences in this experiment, to match the size of the PENN dataset.

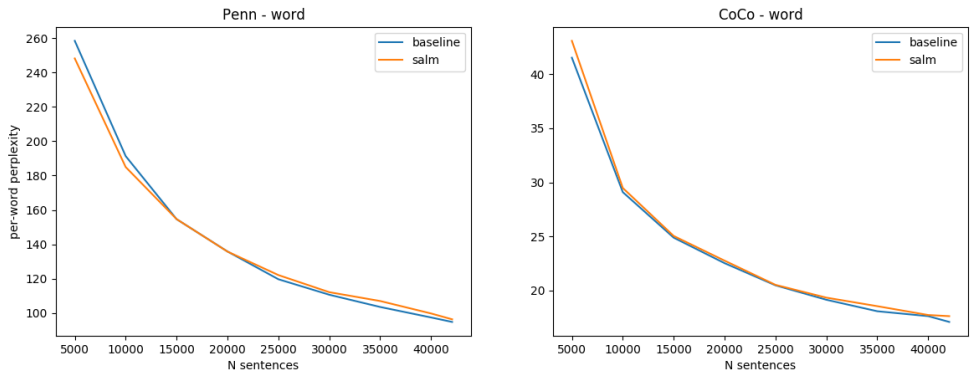

Figure 3: Results of word level experiments

**Text preparation.** We preprocess both corpora by converting the text to lower case and unk-ing all but the most common 10000 words. For the SALM data, we then applied the Stanford parser [7] to the pre-processed training and validation data, and interleaved the words with the PoS tags as per Figure 1. For the PENNdata, we use the supplied validation and test sets. For COCO we use the last 2000 sentences (1000 for validation and 1000 for test).

**Evaluation metric.** In order to perform a direct comparison of SALM and LSTM-LM, we use the standard LM evaluation metric of word-level *model perplexity* [4] on the test sentences (distinct from the validation set) for each dataset. For each word $x_t$ with length $n_t$ one defines the perplexity as:

$$PP(x_t) = 2^{-\frac{1}{n_t}\log_2(p(x_t|\mathbf{x}_{0:t-1}))} \tag{23}$$

**Model variants.** We train four models:

1. a baseline model at the character level in which we use only surface form characters, referred to as LSTM-LM, and our proposed SALM approach which incorporates syntax as token prefixes, instantiated in all three variants discussed in Section 3.1

2. a character level SALM -model using universal PoS tags after application of the parser.

3. a baseline model at the word-level in which we use a dictionary of the words occurring in the corpus.

4. a word-level SALM model, using a dictionary of words and additionally the PoS tags.

**Model training.**

We train the LSTMs with 2 layers, each with $n_h = 256$ units, applying dropout with probability $p_d = 0.2$ to predict the next character in the character string renderings, learning the parameters of the network and initial hidden states $m_{-1}, c_{-1}$ for 40 epochs, choosing the model with best validation perplexity. In all cases we use standard SGD optimization in Pytorch [21], with batches of size 20, annealing the learning rate by dividing by 4 if there was no improvement in validation error. We form batches by concatenating all sentences in the training set and reshaping the sequence into 20 subsequences. These subsequences are then split into blocks of length 35, 70, 217 and 265 for the word level LSTM-LM, SALM and character level LSTM-LM, SALM (respectively) following the principle that the average number of words per block should be invariant between models.

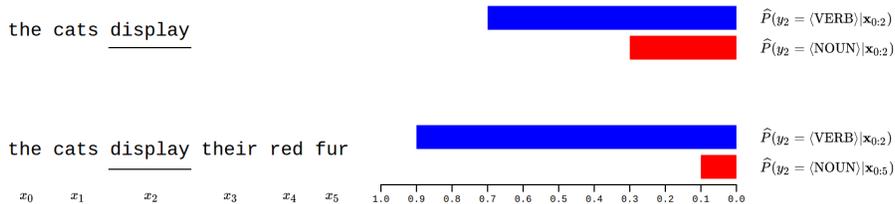

Figure 4: SYNSIR online updating of syntax prefix estimations. In the sequence "the cats display", the word "display" is first assigned a high probability of being a noun (top row). However, as more words are added to the sequence, the probability distribution shifts in favor of "display" to be a verb (bottom row).

We repeat this training setup for $N = 5000, 10000, \ldots, 40000, 42098$, the number of sentences in the training set.

**Model evaluation.** We compare perplexity of the baseline and the perplexity of the SYNSIR estimated probabilities for SALM given by Equation (22). For SYNSIR we set $m = 1000$.

### 4.2 Quantitative Evaluation

The results of the comparative evaluation are displayed in Figures 2 and 3. We observe:

**Syntax-awareness reduces model perplexity in character level models.** For nearly all settings in the character level models, SALM has lower model perplexity than LSTM-LM. These results imply that adding syntax to the model improves its ability to generalize for these models.

**Syntax-awareness doesn't help in word level models.** We see that the word-level SALMs do not show the same level of improvement as the character-level models. This result, for the setup considered, is independent of training sample size. The reason for this difference may be because word-level models are able to learn an implicit model of syntax, meaning that the word-level syntactic annotations are redundant. On the other hand, the explicit word-level markup may allow character level salms to conceptualize grammar and generalize at the word-level more easily.

### 4.3 Show-casing Syntax Sampling

We also qualitatively inspect the resampling method of the proposed SYNSIR algorithm, capable of estimating distributions of syntax prefixes. In this sense, the SALM together with SYNSIR may be considered a shallow syntactic tagger. However, it is also goes beyond parsing in that it may be used to quantify the uncertainty over possible parses given only a sentence fragment. Crucially, it is able to update distributions online as new characters are added to the fragment.

We illustrate this with the sentence "the cats display their red fur", which we pass character-by-character to the algorithm. We visualize the shifts in distribution using a visualization tool, illustrated in Figure 4. As the Figure shows, we observe that given only the first three words ("the cats display"), SYNSIR thinks that "display" is a noun with probabilty $\widehat{P}(y_2|\mathbf{x}_{0:2}) \approx 1/3$. However when the SALM has seen the object of the sentence "fur", the estimate decreases to less than $\widehat{P}(y_2|\mathbf{x}_{0:5}) \approx 0.1$. This shows that the SALM is able to use syntactic information removed several words from a word of interest in order to obtain high-quality predictions and to update its estimates of syntax prefixes given incoming data.

### 4.4 Software and Data Release

We release a python software package for training and applying SALMs using SYNSIR, as well as the preprocessed data and sample generated sentences[1].

## 5 Conclusion and Outlook

We have proposed syntax-aware neural language models (SALMs) and showed that explicitly modeling syntactic structure can substantially improve LM for character level models in terms of model perplexity and data efficiency. We developed a simple approach for rendering syntax as token prefixes and training an LSTM over such sequences. Importantly, we proposed SYNSIR, a sequential Monte Carlo technique which enables us to apply a trained SALM to unparsed text. Our experimental evaluation indicates promise for using this technique to include a richer set of linguistic features into character level language models.

Future work will include incorporating additional linguistic features such as deep syntactic parses and semantic roles [1, 20] into language modeling, as well as further experimentation to improve word-level SALMs.

## 6 Acknowledgments

We thank the members of Zalando research for their valuable feedback and comments.

---

[1] https://github.com/zalandoresearch/salm